\documentclass[runningheads]{llncs}
\usepackage[T1]{fontenc}
\usepackage{graphicx}
\usepackage{hyperref}
\usepackage{url}

\usepackage{subcaption} 
\usepackage{array}      
\usepackage{booktabs}   
\usepackage{multirow}   
\usepackage{tabularx}   
\usepackage{longtable}  
\usepackage{siunitx}    
\usepackage{makecell} % Add this to your LaTeX preamble
\usepackage{algorithm} 
\usepackage{algpseudocode}
\usepackage{amsmath,amssymb,amsfonts}
\usepackage{bbding} % for \Envelope

\usepackage{wrapfig}

\begin{document}
\title{Subject Information Extraction for Novelty Detection with Domain Shifts}
%
%\titlerunning{Abbreviated paper title}
% If the paper title is too long for the running head, you can set
% an abbreviated paper title here
%
% \author{}
%
% \authorrunning{}
% First names are abbreviated in the running head.
% If there are more than two authors, 'et al.' is used.
%
\author{Yangyang Qu \and Dazhi Fu\and Jicong Fan\thanks{Corresponding author}}

% \authorrunning{Y. Qu et al.}

\institute{
School of Data Science\\
The Chinese University of Hong Kong, Shenzhen, China \\
\email{quyang9517@gmail.com, fudazhiaka@gmail.com, fanjicong@cuhk.edu.cn}
}
\maketitle              % typeset the header of the contribution
\begin{abstract}
Unsupervised novelty detection (UND) is vital for applications such as medical diagnosis and cybersecurity. A prevalent assumption in current UND methods is that the normal data for training and testing are drawn from the same domain. This assumption is challenged in practice by domain shift, where a disparity exists between the training and testing domains (e.g., due to different data acquisition pipelines or sites). Domain shift often leads to the incorrect classification of normal data as novel by existing methods. A typical situation is that the normal testing data and normal training data describe the same underlying subject semantics, yet they differ in background/domain conditions. To address this problem, we introduce a novel method that separates subject information from background variation, encapsulating the domain information to enhance detection performance under domain shifts. The proposed method minimizes mutual information between the representations of the subject and background while modeling the background variation using a deep Gaussian mixture model; novelty detection is conducted solely on the subject representations, thus reducing sensitivity to domain variations. Experiments on Multi-background MNIST and Kurcuma demonstrate that our model generalizes effectively to unseen domains and outperforms strong baselines under domain shifts.
\end{abstract}
\section{Introduction}

Novelty detection \cite{markou2003novelty} aims to identify samples that deviate from the training distribution, and is widely used in finance, healthcare, and security. It is closely related to outlier detection \cite{hodge2004survey,fu2026uniod}, anomaly detection \cite{chandola2009anomaly,dai2025autouad}, fault detection \cite{isermann1984process}, and one-class classification \cite{khan2014one}; in particular, unsupervised anomaly detection can be viewed as a special case of novelty detection where (almost) all training data are normal.
In the past decades, numerous novelty detection methods \cite{scholkopf1999support,liu2008isolation,dsvdd,cai2022perturbation,zhang2024deep,xiao2024unsupervised,11071974,cai2025selfdiscriminative} have been proposed. Classical approaches such as kernel density estimation (KDE) \cite{parzen1962estimation} estimate data density and treat low-density samples as novel. Recent deep methods include ALAD \cite{zenati2018adversarially} (GAN-based reconstruction), DeepSVDD \cite{dsvdd} (hypersphere enclosure in latent space), MO-GAAL \cite{liu2019generative} (adversarially generating informative outliers), PLAD \cite{cai2022perturbation} (perturbation learning), and DIF \cite{xu2023deep} (deep isolation forest \cite{liu2008isolation}). More recently, Dai and Fan (2025) \cite{dai2025autouad} introduced AutoUAD, a framework for model selection and hyperparameter tuning in unsupervised anomaly detection, and Fu and Fan (2026) \cite{fu2026uniod} proposed UniOD, a universal model designed for outlier detection across different domains.

In many deployments, normal test data may differ from training data due to acquisition-domain changes (e.g., different scanners/sites or protocols in medical imaging, different cameras/lighting in industrial inspection), while the underlying subject semantics remain unchanged. In this paper, we refer to subject information as the task-relevant semantic factor that should remain stable across domains, and background/domain information as nuisance variations induced by acquisition conditions. Our goal is to detect novelty in terms of subject information rather than background variations.

% \begin{figure}[htbp]
% \begin{center}
% %\framebox[4.0in]{$\;$}
%   \includegraphics[width=0.7\textwidth,trim={0 0 0 0},clip]{illustrate.pdf}
% \end{center}
% \caption{An illustration of the novelty detection task. The training data (left) consists of images of the digit $'0'$ presented in three backgrounds. The testing data (right) includes images of multiple digits $'0-9'$ in seen backgrounds and entirely unseen backgrounds. Although the $'0'$ digits in the test set are normal, some of them are likely to be labelled as novel due to the shift in background.}
% \label{fig:UNSD-TASK}
% \end{figure}

\begin{wrapfigure}{r}{0.5\textwidth}
\vspace{-30pt}
\begin{center}
%\framebox[4.0in]{$\;$}
  \includegraphics[width=1\linewidth,trim={0 0 0 0},clip]{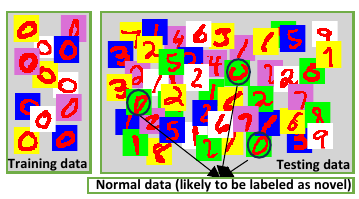}
\end{center}
\vspace{-10pt}
\caption{An illustration of the novelty detection task. The training data (left) consists of images of the digit $'0'$ presented in three backgrounds. The testing data (right) includes images of multiple digits $'0-9'$ in seen backgrounds and entirely unseen backgrounds. Although the $'0'$ digits in the test set are normal, some of them are likely to be labelled as novel due to the shift in background.}
\label{fig:UNSD-TASK}
\vspace{-15pt}
\end{wrapfigure}

A key limitation of many existing unsupervised novelty detection (UND) methods is that they implicitly assume that normal training data and normal testing data come from the same domain. In practice, normal testing data may come from different domains (domain shift), causing models to misclassify normal samples as novel, especially when training and test data describe the same subject but differ in background conditions (Fig.\ref{fig:UNSD-TASK}) \cite{wu2023meta}.

Recent work has started to benchmark and address novelty detection under distribution shift (e.g., GNL) \cite{cao2023anomaly}. Meanwhile, disentanglement is often encouraged by enforcing statistical independence. In this work, we adopt mutual information minimization with a CLUB-style estimator \cite{cheng2020club} to reduce the dependence between subject and background content.

Related ideas such as multi-source augmentation, representation factorization, and removing nuisance information also appear beyond our setting (e.g., cross-modal augmentation \cite{li2025cross}, multi-source modelling \cite{ke2025early}, two-space decomposition \cite{qiu2025convex}, unlearning \cite{li2025multi}, and concept erasure in diffusion/flow models \cite{lu2024mace,gao2024eraseanything}). Beyond images, related directions include video anomaly detection \cite{Lyu_2025,lyu2026forwardconsistencylearninggated} and vision-language/embodied factorization \cite{zeng2025janusvln,zeng2025FSDrive}.

To contextualize our evaluation, we also include representative domain generalization techniques such as ERM \cite{vapnik1991principles} and IRM \cite{arjovsky2019invariant} as transferable baselines by pairing their learned representations with a KDE detector. Unlike these supervised DG methods that typically rely on task labels (and often domain labels), our approach targets unsupervised novelty detection and only assumes the number of background domains in the training set, reducing annotation burden.

To address background (domain) shifts between training and normal test data in UND, we propose Subject-Novelty Detection (SND). SND disentangles subject information from background features in the training data and performs novelty detection solely in the subject representation space, improving robustness to unseen backgrounds. Our main contributions are:
\begin{itemize}
    \item We propose SND, which isolates subject information from background variations to enable robust novelty detection under significant domain shifts.
    \item SND requires only the number of domains in the training data (without per-sample domain labels), making it practical for low-annotation settings.
    \item SND outperforms strong novelty detection and domain-shift baselines in the experiments.
\end{itemize}

\section{Unsupervised Subject Novelty Detection}
\label{headings}

\subsection{Problem Formulation}

To be precise, suppose we have a training dataset consisting of $N$ images, denoted as $ \mathcal{D} = \{\mathbf{{x}}_1,\mathbf{{x}}_2,\ldots,\mathbf{{x}}_N\}$, in which each $\mathbf{x}_i\in\mathbb{R}^{C\times H\times W}$ has a background $b_i$ chosen from a set of $K$ different backgrounds $\mathcal{B}=\{B_1,B_2,\ldots,B_K\}$ and all or most of the $N$ samples are normal. Notably, although the total number of backgrounds $K$ is known, the specific background type of each image is unknown. This setting is practical since data or images collected often come from different backgrounds (or domains more generally), and labeling the backgrounds is costly. We consider a test set $\mathcal{D}'= \{\mathbf{x}{'}_1, \mathbf{{x}}{'}_2, \ldots, \mathbf{x}{'}_M\}$, where the background $b_i{'}$ of each $\mathbf{x}{'}_i$ is chosen from a larger set $\tilde{\mathcal{B}}=\{B_1,B_2,\ldots,B_K,B_{K+1},\ldots,B_{K+K'}\}$. Note that $B_{K+1},\ldots,B_{K+K'}$ are actually new backgrounds different from $B_1,B_2,\ldots,B_K$ and $K'$ is unknown.
Our goal is to learn a model from $\mathcal{D}$ to determine whether a new sample from $\mathcal{D}'$ is a novel sample in terms of the subject information rather than the background information. This is a nontrivial task because the domain (or distribution) of normal data changed. 

An example of the task is shown in Fig.\ref{fig:UNSD-TASK}, where the training set contains images of digit $'0'$ with $3$ different colored backgrounds (white, yellow, blue), and the testing set contains images of digits $'0-9'$ with an additional unseen background (green). To evaluate performance under background shifts, the background of digit $'0'$ (normal testing data) is set to a completely unseen green background, while the backgrounds of digits $'1-9'$ (testing novel data) may appear in both seen and unseen backgrounds. Our aim is to identify digits $'1-9'$ as novel samples which contain different subject information, while treating digits $'0'$ as normal samples which differ from training data only in background information. 

Classical ND tasks only consider the distribution difference between the training data and the testing novel data. Our tasks consider not only the distribution difference mentioned before but also the background (domain) shift between training data and testing normal data, which leads to a distribution difference between them. Therefore, unsupervised subject novelty detection is more challenging, as classical ND methods often mislabel normal samples with new backgrounds as novel, leading to high false positive rates.

\begin{figure}[t]
\begin{center}
%\framebox[4.0in]{$\;$}
  \includegraphics[width=0.75\textwidth,trim={0 20 0 0},clip]{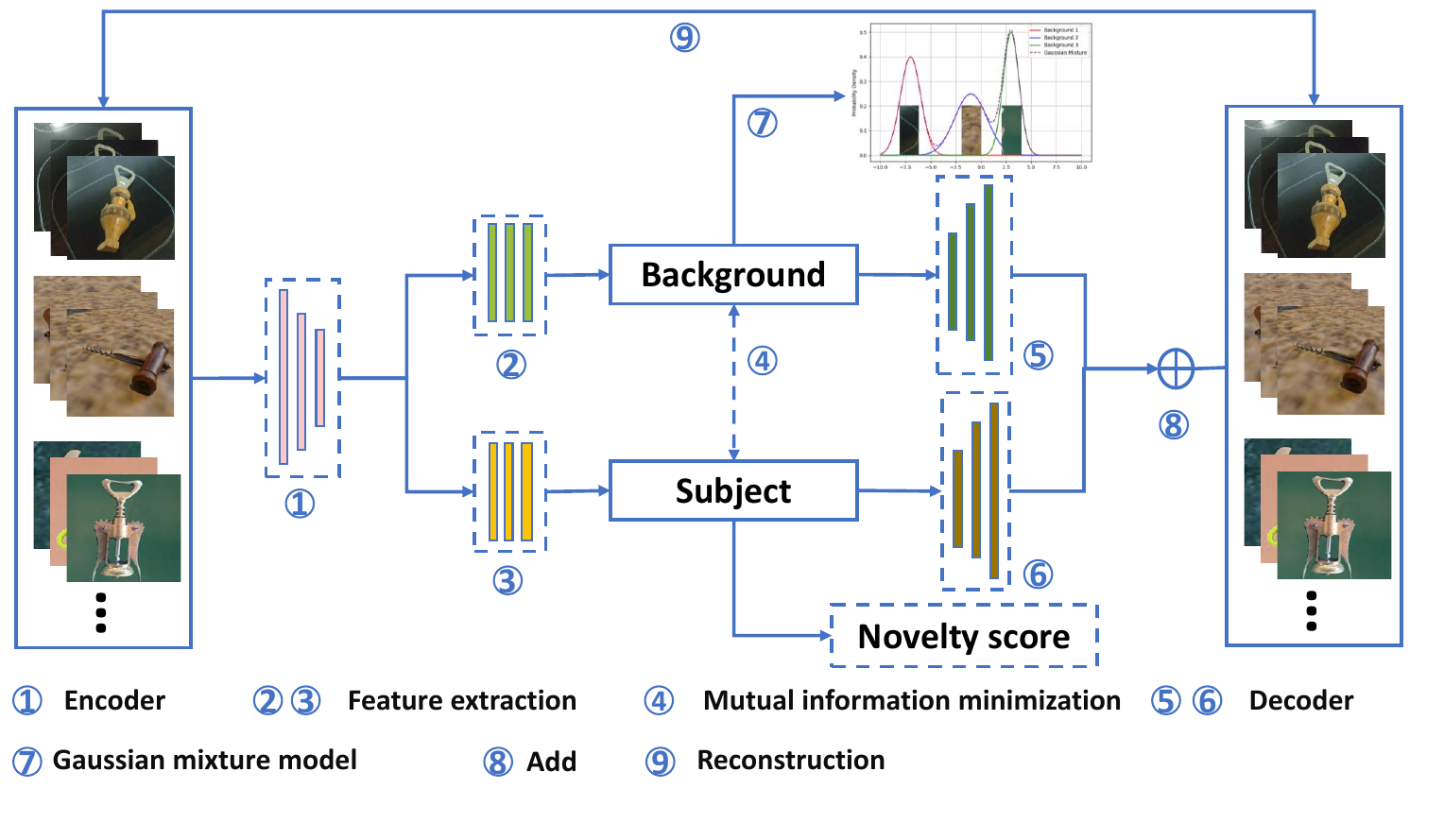} % Replace with your image file name
\end{center}
\vspace{-15pt}
\caption{An overview of the proposed SND model.}
\label{fig:structure}
\vspace{-15pt}
\end{figure}

\subsection{Proposed Model}

We aim to address the challenge of isolating subject information from varying backgrounds for improved novelty detection. One key point is to learn representations that separate subject and background features in an unsupervised manner, allowing the model to detect novel subject information despite shifts in background domains. The overall process of our method is illustrated in Figure~\ref{fig:structure}.

The process begins with a feature extraction network \( G_{\theta_f}: \mathbb{R}^{C\times H\times W}\rightarrow \mathbb{R}^d \) with parameters $\theta_f$, which processes the input image \( \mathbf{x} \) and generates a feature representation
\( \mathbf{z}_f \), i.e.,
\begin{equation}
\label{eq:zf}
\mathbf{z}_f = G_{\theta_f}(\mathbf{x}).
\end{equation}
This representation is then decomposed into two distinct components, a subject feature vector \( \mathbf{z}_s \) and a background feature vector \( \mathbf{z}_b \), through two neural networks \( F_{\theta_s}:\mathbb{R}^d\rightarrow \mathbb{R}^d  \) and
\( F_{\theta_b}:\mathbb{R}^d\rightarrow \mathbb{R}^d \), i.e.,
\begin{equation}
\label{eq:zs_zb}
\mathbf{z}_s = F_{\theta_s}(\mathbf{z}_f), \quad
\mathbf{z}_b = F_{\theta_b}(\mathbf{z}_f).
\end{equation}

It is nontrivial to guarantee that \( \mathbf{z}_s \) and \( \mathbf{z}_b \) exclusively represent the subject and background
information, respectively. Following the insights of CLUB \cite{cheng2020club}, we propose to minimize the mutual information
\( I(\mathbf{z}_s; \mathbf{z}_b) \) between \( \mathbf{z}_s \) and \( \mathbf{z}_b \), which encourages the two parts to be statistically independent. The mutual information is estimated using a neural network \( \xi_{\theta_m} \) based on the following formulation:
\begin{align}
\label{eq:mi}
\hat{I}_{\text{MI}}(\mathbf{z}_s; \mathbf{z}_b)
= \frac{1}{N} \sum_{i=1}^{N} \left[
   \log \left(\xi_{\theta_m}(\mathbf{z}_b^{(i)} \mid \mathbf{z}_s^{(i)})\right)
   - \frac{1}{N} \sum_{j=1}^{N}
       \log \left(\xi_{\theta_m}(\mathbf{z}_b^{(j)} \mid \mathbf{z}_s^{(i)})\right)
 \right].
\end{align}
In practice, we parameterize $\xi_{\theta_m}(\mathbf{z}_b \mid \mathbf{z}_s)$ as a Gaussian distribution and adopt an alternating training scheme: we first update $\theta_m$ by maximizing $\hat{I}_{\text{MI}}$ with encoders fixed, and then update the encoder parameters $G_{\theta_f}, F_{\theta_s}, F_{\theta_b}$ by minimizing $\hat{I}_{\text{MI}}$ while detaching the gradients through $\theta_m$.

It should be noted that making $\mathbf{z}_s$ and $\mathbf{z}_b$ independent alone cannot ensure that $\mathbf{z}_s$ is composed of subject information and $\mathbf{z}_b$ is composed of background information; they may swap roles. To address this, we assume that the number of background types is $K$ and that $K$ is different from the number of potential clusters in the subject information. Under this assumption, we can distinguish between subject and background information. Specifically, inspired by \cite{zong2018deep}, we use a deep Gaussian mixture model (GMM) with $K$ components to model \( \mathbf{z}_b \). A neural network $S_{\theta_g}:\mathbb{R}^d\rightarrow \mathbb{R}^K$ projects $\mathbf{z}_b^{(i)}$ to soft membership predictions
$\hat{\boldsymbol{\gamma}}^i \in \mathbb{R}^K$ for each mixture component:
\begin{equation}
\label{eq:gamma}
\hat{\boldsymbol{\gamma}}^i = \operatorname{softmax}\!\big(S_{\theta_g}(\mathbf{z}_b^{(i)})\big).
\end{equation}
This modeling encourages \( \mathbf{z}_b \) to capture $K$ clusters, making it different from the subject information. Denoting \( \hat{\pi}_k \) as the weight of the \( k \)-th Gaussian component and \( \boldsymbol{\hat{\mu}}_k\in\mathbb{R}^d \) as the mean, we estimate the mixture parameters from a mini-batch $\{\mathbf{z}_b^{(i)}\}_{i=t_1}^{t_L}$ of size $L$ as
% \begin{equation}
% \label{eq:gmm-params}
% \begin{aligned}
% \hat{\pi}_k
% &= \frac{1}{L} \sum_{i=t_1}^{t_L} \hat{\gamma}_k^i, \quad
% \hat{\boldsymbol{{\mu}}}_k
% = \frac{\sum_{i=t_1}^{t_L} \hat{\gamma}_k^i \mathbf{z}_b^{(i)}}
%        {\sum_{i=t_1}^{t_L} \hat{\gamma}_k^i}, \\
% \hat{\boldsymbol{{\Sigma}}}_k
% &= \frac{\sum_{i=t_1}^{t_L} \hat{\gamma}_k^i
%          (\mathbf{z}_b^{(i)} - \hat{\boldsymbol{{\mu}}}_k)
%          (\mathbf{z}_b^{(i)} - \hat{\boldsymbol{{\mu}}}_k)^\top}
%         {\sum_{i=t_1}^{t_L}\hat{\gamma}_k^i}.
% \end{aligned}
% \end{equation}
\begin{equation}
\label{eq:gmm-params}
\begin{aligned}
\hat{\pi}_k
&= \frac{1}{L} \sum_{i=t_1}^{t_L} \hat{\gamma}_k^i, \quad
\hat{\boldsymbol{\mu}}_k
= \frac{\sum_{i=t_1}^{t_L} \hat{\gamma}_k^i \mathbf{z}_b^{(i)}}
       {\sum_{i=t_1}^{t_L} \hat{\gamma}_k^i}, \\
\hat{\boldsymbol{\sigma}}_k^2
&= \frac{\sum_{i=t_1}^{t_L} \hat{\gamma}_k^i
         \big(\mathbf{z}_b^{(i)} - \hat{\boldsymbol{\mu}}_k\big)\odot
         \big(\mathbf{z}_b^{(i)} - \hat{\boldsymbol{\mu}}_k\big)}
        {\sum_{i=t_1}^{t_L}\hat{\gamma}_k^i}, \\
\mathbf{s}_k
&= \log\!\big(\hat{\boldsymbol{\sigma}}_k^2 + \varepsilon\big), \quad
\hat{\boldsymbol{\Sigma}}_k = \operatorname{diag}\!\big(\exp(\mathbf{s}_k)\big).
\end{aligned}
\end{equation}
where $\odot$ denotes element-wise multiplication, and $\log(\cdot)$ and $\exp(\cdot)$ are applied element-wise. $\varepsilon>0$ is a small constant for numerical stability.

Given a randomly sampled batch of data $\{\mathbf{x}_{i}\}_{i=t_1}^{t_L} \subseteq \mathcal{D}$ and their
background feature vectors $\{\mathbf{z}_b^{(i)}\}_{i=t_1}^{t_L}$, we define the following
background energy function:
\begin{equation}
\label{eq:gmm-energy}
\begin{aligned}
E(\mathbf{z}_b^{(i)})
= -\log \Bigg(
  \sum_{k=1}^{K} \hat{\pi}_k (2\pi)^{-d/2}
  \big|\boldsymbol{\hat{\Sigma}}_k\big|^{-1/2}
  \exp \Big(
    -\tfrac{1}{2}
    (\mathbf{z}_b^{(i)} - \boldsymbol{\hat{\mu}}_k)^\top
    \boldsymbol{\hat{\Sigma}}_k^{-1}
    (\mathbf{z}_b^{(i)} - \boldsymbol{\hat{\mu}}_k)
  \Big)
\Bigg).
\end{aligned}
\end{equation}
The identification of \( \mathbf{z}_b \) as the background latent variable together with its enforced independence from \( \mathbf{z}_s \) encourages \( \mathbf{z}_s \) to capture the subject information.

Nevertheless, we still need to ensure that $\mathbf{z}_s$ and $\mathbf{z}_b$ preserve the original information of the input
$\mathbf{x}$. This is achieved by reconstructing $\mathbf{x}$ from the two latents. Specifically, we feed $\mathbf{z}_s$ and $\mathbf{z}_b$ into two decoders
$H_{\theta_s'}: \mathbb{R}^d\rightarrow \mathbb{R}^{C\times H\times W}$
and
$H_{\theta_b'}: \mathbb{R}^d\rightarrow \mathbb{R}^{C\times H\times W}$, i.e.:
\begin{equation}
\mathbf{x}_s = H_{\theta_s'}(\mathbf{z}_s), \quad
\mathbf{x}_b = H_{\theta_b'}(\mathbf{z}_b),
\end{equation}
and combine their outputs as the reconstruction of $\mathbf{x}$:
\begin{equation}
\hat{\mathbf{x}} = \mathbf{x}_s + \mathbf{x}_b.
\end{equation}

By isolating subject and background information in this way, our method focuses on detecting novelty in the subject information, even when there is significant variation in the background. This feature decomposition and reconstruction mechanism ensures robustness to background changes and facilitates accurate novelty detection. Note that the mixture parameters $\{\hat{\pi}_k,\hat{\boldsymbol{\mu}}_k,\hat{\boldsymbol{\Sigma}}_k\}_{k=1}^K$ are estimated from each mini-batch using the soft responsibilities $\hat{\gamma}_k^i$ in \eqref{eq:gmm-params}, and gradients are back-propagated through both the responsibilities and the batch statistics during training.
\subsection{Training and Evaluation}

Here, we summarize the entire process of the proposed method. Due to the mutual information estimation and GMM parts, we can ensure that both \( \mathbf{z}_s \) and \( \mathbf{z}_b \) contain necessary information about the subject and background, respectively, without worrying that one has learned most of the information while the other has not learned anything. The loss function \( L_{\text{rec}}(\mathbf{x}, \hat{\mathbf{x}}) \) represents the reconstruction error between the original image \( \mathbf{x} \) and the reconstructed output image \( \hat{\mathbf{x}} \), which is expressed as
\begin{equation}
\label{loss_l2}
L_{\text{rec}}(\mathbf{x}, \hat{\mathbf{x}}) = \left\| \mathbf{x} - \hat{\mathbf{x}} \right\|_2^2.
\end{equation}
We calculate the weighted sum of all the loss terms to obtain the total loss for the proposed model:
\begin{equation}
\label{eq-loss_all}
L_{\text{total}} =  L_{\text{rec}}(\mathbf{x}, \hat{\mathbf{x}}) + \omega_1 E(\mathbf{z}_b) + \omega_2\hat{I}_{\text{MI}}(\mathbf{z}_s; \mathbf{z}_b),
\end{equation}
where \( \omega_1 \) and \( \omega_2 \) are non-negative hyperparameters, and the parameters to learn are $\{\theta_f,\theta_s,\theta_b,\theta_m,\theta_g, \theta_s',\theta_b'\}$.

After our model is well-trained, we can use KDE \cite{parzen1962estimation}, a simple yet effective method, to perform novelty detection. Specifically, we denote the subject feature vectors of the training set as 
$\mathcal{D}_{s} = \{ \mathbf{z}_s^{(1)}, \mathbf{z}_s^{(2)}, \ldots, \mathbf{z}_s^{(N)} \} = \{ F_{\theta_s}(G_{\theta_f}(\mathbf{x})) : \mathbf{x} \in \mathcal{D} \}$.
Given a test sample \( \mathbf{x}_{\text{new}} \), its subject feature vector is \( \mathbf{z}_s^{\text{new}} = F_{\theta_s}(G_{\theta_f}(\mathbf{x}_{\text{new}})) \).  The novelty score (NS) of \( \mathbf{x}_{\text{new}} \) is given by the negative density of $\mathbf{z}_s^{\text{new}}$, i.e.,
\begin{equation}
\label{eq-score}
\text{NS}(\mathbf{x}_{\text{new}}) = -\hat{p}(\mathbf{z}_s^{\text{new}}) = -\frac{1}{N (2\pi h^2)^{d/2}} \sum_{i=1}^{N} \exp\left( -\frac{\| \mathbf{z}_s^{\text{new}} - \mathbf{z}_s^{(i)} \|^2}{2 h^2} \right)
\end{equation}
where \( h \) is the bandwidth parameter controlling the smoothness of the estimated density, and \( d \) is the dimensionality of \( \mathbf{z}_s \).
A higher novelty score $\text{NS}(\mathbf{x}_{\text{new}})$ indicates that the subject of $\mathbf{x}_{\text{new}}$ has a lower likelihood of belonging to the distribution of subjects in the training data $\mathcal{D}$.

In general, the proposed method learns comprehensive subject features \( \mathbf{z}_s \) and background features \( \mathbf{z}_b \) using our objective function defined in \eqref{eq-loss_all}, which includes the weighted sum of reconstruction loss, energy of \( \mathbf{z}_b \), and mutual information between \( \mathbf{z}_s \) and \( \mathbf{z}_b \). For novelty detection, KDE fitted on the training set is applied to the subject feature \( \mathbf{z}_s \) of the test sample, and the novelty score for a test sample is determined using \eqref{eq-score}.

\section{Experiments}

In this section, we benchmark our method and various baselines using numerical experiments on several challenging and widely used datasets. We adopt a leave-one-class-out protocol: for each run, we select 9 out of 10 classes as normal for training and treat the remaining class as novel for testing; we report per-class results and the average across all 10 splits. For testing, images from a different unseen background are used. 

\subsection{Datasets, Compared Methods, and Evaluation Metrics}

We evaluated the proposed method on two challenging datasets: Multi-background MNIST and Kurcuma \cite{rosello2023kurcuma}. To address the limitations in variability in the original MNIST datasets \cite{lecun1998gradient}, we introduced domain shifts by altering background colors. For the Multi-background MNIST dataset, the model was trained using ‘blue’, ‘yellow’, and ‘white’ backgrounds and tested on a previously unseen ‘green’ background. These setups evaluated the model's generalization to unseen domains.
Additionally, we use Kurcuma, a benchmark for kitchen-utensil recognition in home-assistance robotics that comprises seven corpora/domains with diverse appearance shifts (e.g., synthetic, clipart, and real-image settings) \cite{rosello2023kurcuma}.

\begin{table}[t]\small
\centering
\caption{Average AUROCs (\%) in novelty detection on Multi-background MNIST. In each case, the best result is marked in bold.}
\label{tab:mnist_auc}
\begin{tabularx}{\textwidth}{l*{10}{>{\centering\arraybackslash}X}c}
\toprule
Method & 0 & 1 & 2 & 3 & 4 & 5 & 6 & 7 & 8 & 9 & Average\\
\midrule
COPOD     & 62.82 & 70.33 & 63.77 & 64.41 & 65.80 & 64.33 & 64.44 & 66.05 & 63.81 & 65.95 & 65.17   \\
SUOD      & 64.52 & 67.42 & 65.79 & 67.72 & 70.17 & 69.37 & 65.46 & 67.24 & 65.10 & 67.70 & 67.05   \\
MO\_GAAL  & 61.41 & 72.20 & 69.40 & 77.73 & 65.74 & 58.00 & 71.48 & 71.23 & 73.13 & 73.63 & 69.40   \\
DeepSVDD  & 63.92 & 58.84 & \textbf{72.46} & 56.69 & 48.24 & \textbf{88.63} & 66.82 & 76.00 & 62.02 & 62.95 & 65.66   \\
ALAD      & 27.97 & 29.07 & 7.58  & 17.22 & 8.85  & 25.06 & 13.84 & 9.77  & 22.37 & 16.93 & 17.87   \\
ECOD      & 57.29 & 61.46 & 60.37 & 60.41 & 62.13 & 61.19 & 60.22 & 61.29 & 59.67 & 61.53 & 60.56   \\
INNE      & 61.23 & 57.83 & 63.72 & 63.15 & 61.80 & 66.64 & 62.72 & 65.09 & 58.84 & 61.37 & 62.24   \\
AnoGAN    & 4.86  & 0.36  & 32.28 & 52.98 & 52.43 & 43.97 & 9.63  & 33.07 & 22.85 & 36.38 & 28.88   \\
\midrule
ERM       & 36.42 & 95.36 & 42.00 & 42.25 & 38.25 & 40.29 & 51.65 & 51.96 & 48.00 & 40.54 & 48.67   \\
IRM       & 35.65 & 96.32 & 40.41 & 37.54 & 38.47 & 37.09 & 47.72 & 63.56 & 47.82 & 41.29 & 48.59   \\
GNL       & 61.47 & 93.07 & 50.04 & 82.73 & 63.20 & 54.30 & 60.23 & 68.56 & 60.58 & 56.51 & 65.07   \\
\midrule
SND       & \textbf{85.74}      & \textbf{97.68} & 71.35 & \textbf{84.40} & \textbf{75.55} & 74.59 & \textbf{90.39} & \textbf{85.09} & \textbf{80.24} & \textbf{74.08} & \textbf{82.27}   \\
\bottomrule
\end{tabularx}
\end{table}

% \textbf{Compared Methods and Evaluation Metrics} 

We conducted an extensive performance evaluation by comparing our model against a wide range of recent state-of-the-art novelty detection methods. We include both classical and deep novelty detection baselines; all methods are implemented using PyOD defaults when applicable. Methods compared includes  AnoGAN \cite{schlegl2017unsupervised}, DeepSVDD \cite{dsvdd}, XGBOD \cite{zhao2018xgbod}, ALAD \cite{zenati2018adversarially}, INNE \cite{bandaragoda2018isolation}, MO-GAAL \cite{liu2019generative}, COPOD \cite{li2020copod}, ROD \cite{almardeny2020novel}, SUOD \cite{zhao2021suod}, and ECOD \cite{li2022ecod}. The hyperparameters for the methods were set according to the default settings provided by the PyOD\cite{zhao2019pyod}. 
Furthermore, we evaluated our approach against GNL, a recently proposed method for novelty detection across domain transformations \cite{cao2023anomaly}. We also compared our method with two key domain adaptation techniques, ERM and IRM \cite{arjovsky2019invariant}, both followed by a KDE step for novelty detection. This allowed us to evaluate our model's effectiveness in handling domain shifts and identifying novel data in unseen environments.

We employed two common metrics to evaluate the performance of novelty detection: (i) Area Under the Receiver Operating Characteristic curve (AUROC), which can be interpreted as the probability that a positive sample has a higher discriminative score than a negative sample; and (ii) Area Under the Precision-Recall curve (AUPRC), an ideal metric for adjusting extreme differences between positive and negative base rates.

\begin{table}[t!]
\caption{Average AUROC (\%) on Kurcuma across seven target domains. Columns 0--8 correspond to the held-out novel class (9 categories in total).}
\label{tab:kurcuma_all_auc}
\begin{tabularx}{\textwidth}{l*{11}{>{\centering\arraybackslash}X}}
\toprule
Method & 0 & 1 & 2 & 3 & 4 & 5 & 6 & 7 & 8 & Average \\
\midrule
ALAD     & 47.60 & 46.97 & 48.78 & 55.30 & 47.37 & 49.41 & 48.87 & 50.12 & 50.27 & 49.41   \\
COPOD    & 45.51 & 43.85 & 55.37 & 50.76 & 50.99 & 48.32 & 54.48 & 47.18 & 51.75 & 49.80   \\
DeepSVDD & 48.79 & 46.12 & 49.64 & 50.87 & 51.27 & 48.31 & 52.12 & 50.55 & 50.22 & 49.77   \\
ECOD     & 45.97 & 43.95 & 55.23 & 50.63 & 49.10 & 49.31 & 54.56 & 47.37 & 49.52 & 49.51   \\
INNE     & 47.09 & 41.52 & 58.66 & 53.47 & 45.48 & 42.89 & 59.00 & 47.60 & 52.87 & 49.84   \\
AnoGAN   & 47.39 & 52.70 & 47.23 & 50.48 & 57.89 & 47.59 & 49.42 & 46.73 & 48.82 & 49.81   \\
\midrule
ERM      & 55.37 & 45.96 & 47.05 & 50.42 & 51.17 & 48.36 & 47.08 & 55.49 & 50.18 & 50.12   \\
IRM      & 53.71 & 47.71 & 47.23 & 50.19 & 51.90 & 48.76 & 50.25 & 53.45 & 43.56 & 49.64   \\
GNL      & 49.29 & 41.46 & \textbf{77.32} & 65.75 & 48.81 & \textbf{62.32} & \textbf{71.97} & 61.11 & \textbf{71.67} & 61.08   \\
\midrule
SND      & \textbf{72.70} & \textbf{71.56} & 74.13 & \textbf{65.85} & \textbf{78.67} & 59.98 & 64.18 & \textbf{71.72} & 70.85 & \textbf{69.96}   \\
\bottomrule
\end{tabularx}
\end{table}
\subsection{Results and Discussion}
In this section, we evaluate and analyze the performance of our method compared to recent novelty detection baseline methods across the mentioned datasets. SND consistently demonstrates superior performance in the extensive experiments.

In Table \ref{tab:mnist_auc}, we compare the performance of various methods on novelty detection using the Multi-background MNIST dataset, focusing on AUROC scores. Our proposed method, SND, achieves the highest average AUROC of 82.27\%, outperforming baseline methods like COPOD (65.17\%) and SUOD (67.05\%). ERM and IRM, two domain adaptation techniques followed by KDE for novelty detection, perform significantly lower with averages of 48.67\% and 48.59\%, respectively. Notably, SND excels in digits such as 0 (85.74\%) and 1 (97.68\%), demonstrating superior generalization across different digits. Due to the space limitations, we will not report the results of AUPRC, though our method still outperformed the baselines in terms of AUPRC.

\begin{table}[t]
\caption{Average AUPRC (\%) on Kurcuma across seven target domains. Columns 0--8 correspond to the held-out novel class (9 categories in total).}
\label{tab:kurcuma_prc}
\begin{tabularx}{\textwidth}{l*{11}{>{\centering\arraybackslash}X}}
\toprule
Method & 0 & 1 & 2 & 3 & 4 & 5 & 6 & 7 & 8 & Average \\
\midrule
ALAD     & 91.82 & 94.30 & 90.33 & 80.77 & 95.29 & 82.79 & 78.94 & 93.48 & 93.08 & 88.98   \\
COPOD    & 91.31 & 94.45 & 91.74 & 79.00 & 94.96 & 82.72 & 81.30 & 93.22 & 93.24 & 89.10   \\
DeepSVDD & 92.14 & 94.15 & 90.22 & 79.04 & 95.52 & 83.07 & 81.11 & 93.72 & 92.96 & 89.10   \\
ECOD     & 91.55 & 94.26 & 91.52 & 79.04 & 94.92 & 83.30 & 80.90 & 93.18 & 93.19 & 89.10   \\
INNE     & 91.80 & 93.80 & 92.77 & 80.09 & 94.74 & 79.96 & 83.45 & 93.29 & 93.90 & 89.31   \\
AnoGAN   & 92.36 & 95.32 & 89.04 & 79.28 & 96.49 & 82.30 & 80.20 & 93.05 & 93.29 & 89.04   \\
\midrule
ERM      & 93.41 & 94.52 & 89.41 & 79.38 & 95.86 & 82.08 & 77.95 & 94.62 & 92.93 & 88.91   \\
IRM      & 93.45 & 94.79 & 88.99 & 79.95 & 95.81 & 83.22 & 77.84 & 94.42 & 92.96 & 89.05   \\
GNL      & 91.88 & 93.97 & \textbf{96.47} & 85.65 & 95.08 & \textbf{88.77} & \textbf{89.14} & 95.23 & \textbf{96.52} & 92.52   \\
\midrule
SND      & \textbf{96.00} & \textbf{97.67} & 96.43 & \textbf{87.08} & \textbf{98.36} & 87.55 & 88.73 & \textbf{96.82} & 96.38 & \textbf{93.89} \\  
\bottomrule
\end{tabularx}
\vspace{-10pt}
\end{table}

In Table \ref{tab:kurcuma_all_auc}, we show the AUROC results on the Kurcuma dataset, which includes seven distinct scenarios: SYNTHETIC, AKUD, CLIPART, EKUD, EKUD-M1, EKUD-M2, and EKUD-M3. Each scenario corresponds to a specific category: 0 for bottle opener, 1 for can opener, 2 for fork, 3 for knife, 4 for pizza cutter, 5 for spatula, 6 for spoon, 7 for tongs, and 8 for whisk. SND achieves the best average AUROC of 69.96\%, outperforming other methods across most categories, including 0 (72.70\%) and 4 (78.67\%). GNL shows strong results in category 2 (77.32\%) but falls short overall with an average of 61.08\%. ERM and IRM trail behind with averages of 50.12\% and 49.64\%. Table \ref{tab:kurcuma_prc} further shows that SND achieves the best average AUPRC.

\begin{figure}[t]
    \centering
    \begin{subfigure}[b]{0.32\linewidth}
        \centering
        \includegraphics[width=\linewidth]{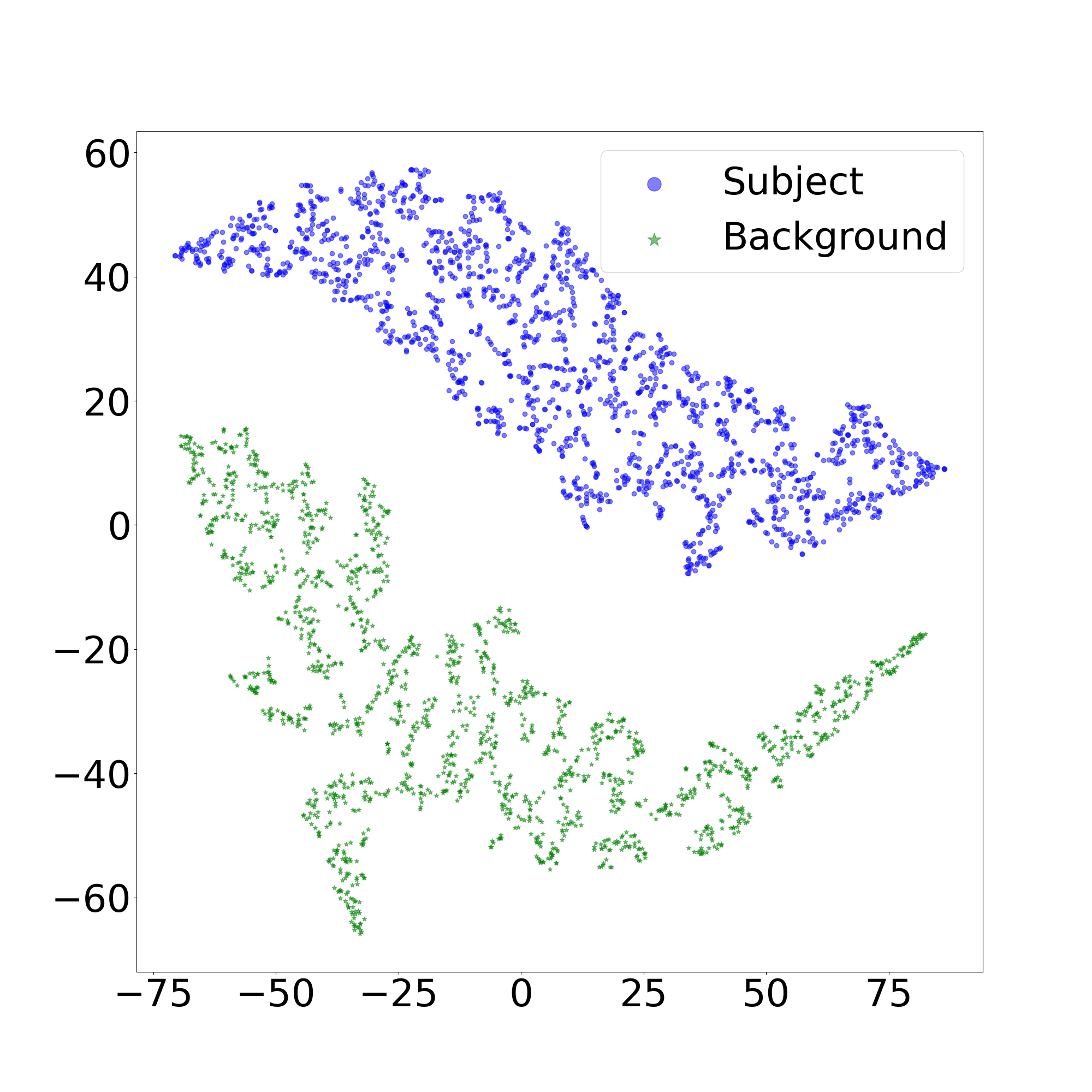}
        \caption{Subject vs.\ Background}
        \label{fig:tsne_subject_background}
    \end{subfigure}
    \hfill 
    \begin{subfigure}[b]{0.32\linewidth}
        \centering
        \includegraphics[width=\linewidth]{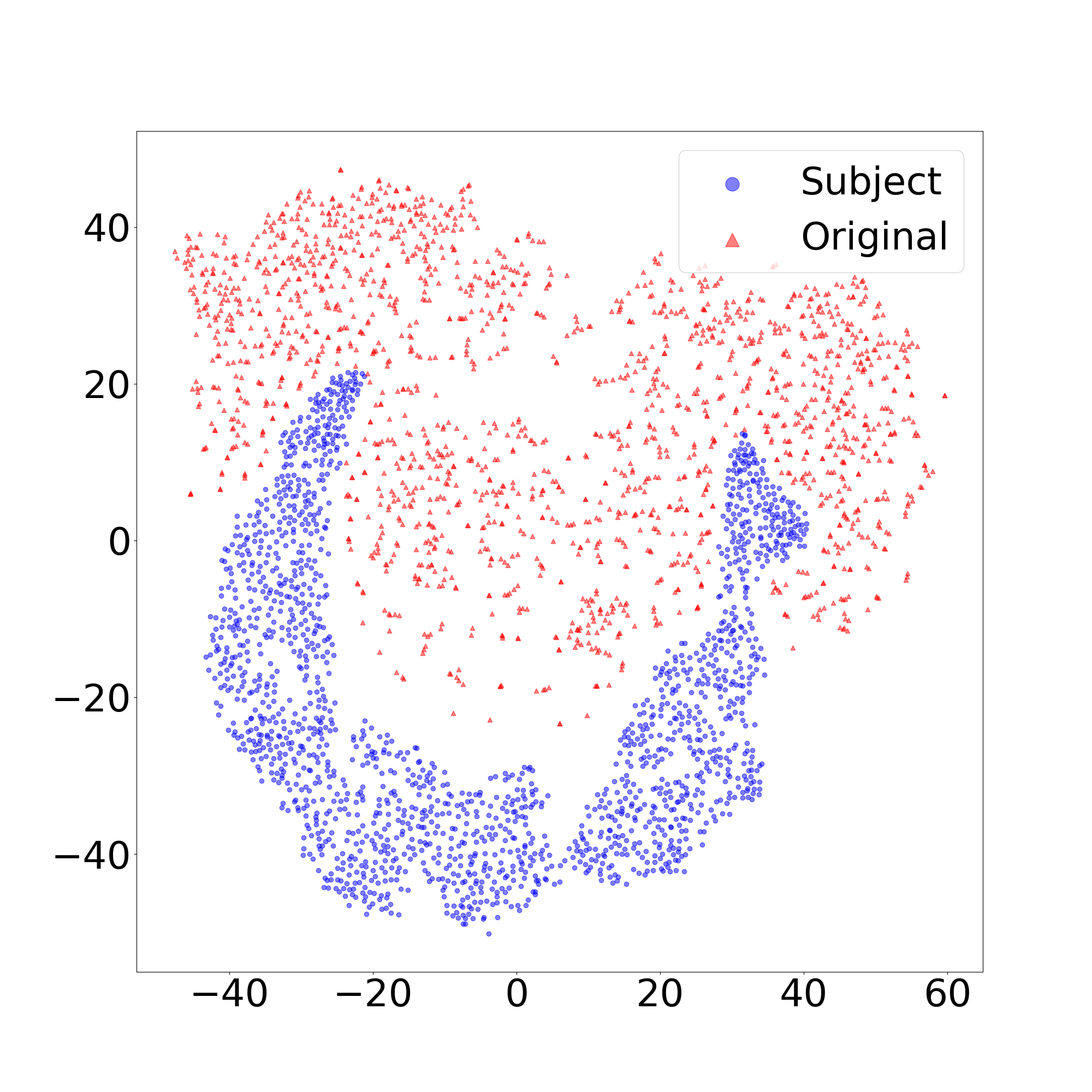}
        \caption{Subject vs.\ Original}
        \label{fig:tsne_subject_original}
    \end{subfigure}
    \hfill 
    \begin{subfigure}[b]{0.32\linewidth}
        \centering
        \includegraphics[width=\linewidth]{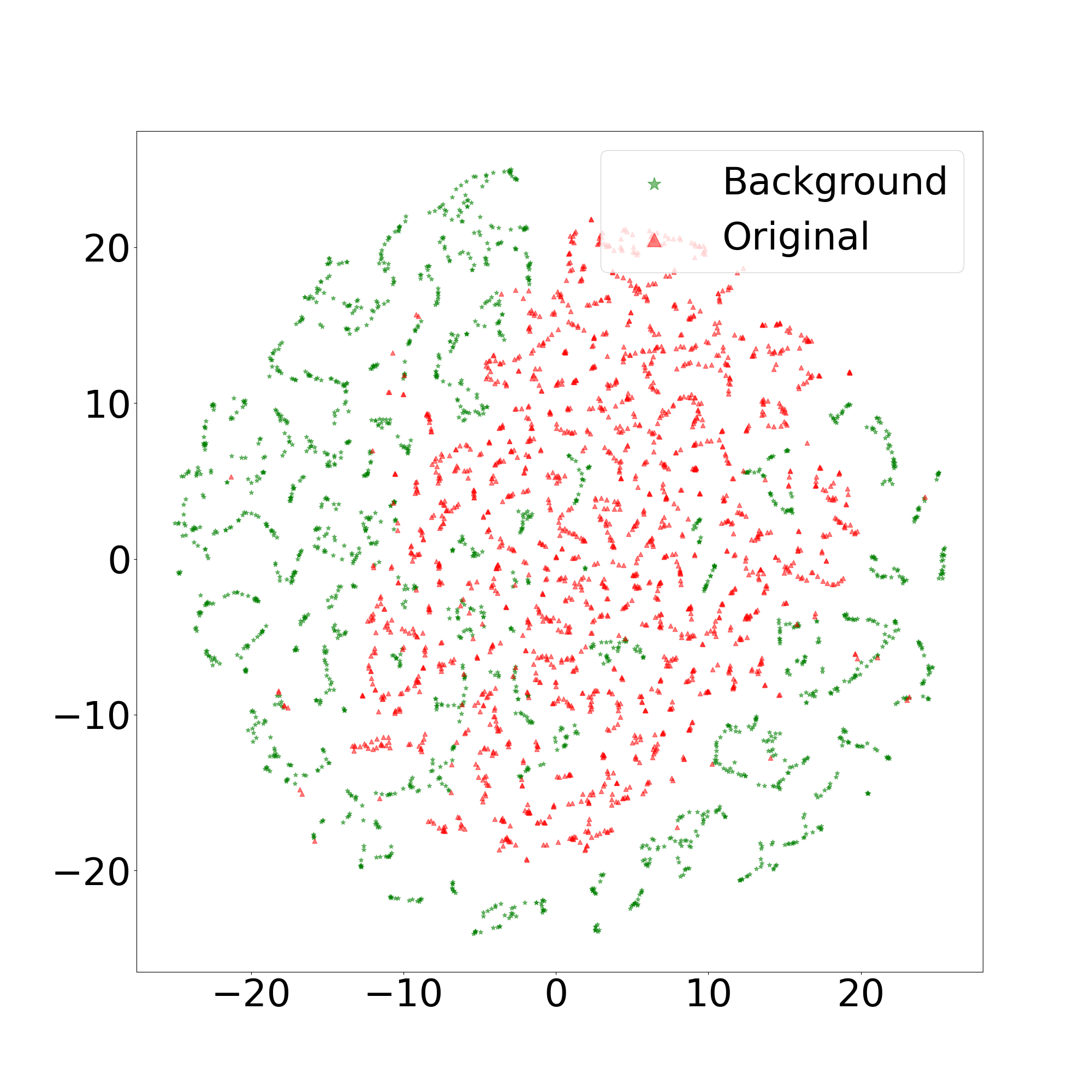}
        \caption{Background vs.\ Original}
        \label{fig:tsne_background_original}
    \end{subfigure}
    \caption{t-SNE visualization of the learned features: (a) Subject vs.\ Background, (b) Subject vs.\ Original, and (c) Background vs.\ Original. We chose the class of “0” in the Multi-background MNIST to provide the visualization result.}
    \label{fig:tsne_results}
\end{figure}

% In addition, we use t-SNE \cite{van2008visualizing} to quantitatively evaluate SND's performance under domain shifts. Figs.\ref{fig:tsne_subject_background}, \ref{fig:tsne_subject_original}, and  \ref{fig:tsne_background_original} demonstrate the model’s capacity to isolate subject features across varying scenarios, as shown through t-SNE visualizations. Specifically, Fig.\ref{fig:tsne_subject_background} shows the t-SNE projection of subject and background features, revealing distinct clusters that highlight effective feature separation. Fig.\ref{fig:tsne_subject_original} highlights the t-SNE results for subject and original image features, further confirming that the model retains essential subject information while discarding irrelevant background details. Finally, in Fig.\ref{fig:tsne_background_original}, the comparison of background features with those of the original images reveals the model's capacity to distinguish between background elements and the overall image characteristics.

Additionally, we use t-SNE \cite{van2008visualizing} to qualitatively evaluate SND’s feature representation under domain shift. As shown in Figs.~\ref{fig:tsne_subject_background}, \ref{fig:tsne_subject_original}, and \ref{fig:tsne_background_original}, the visualizations confirm the model's ability to isolate subject features by forming distinct clusters for different domains, effectively separating subject from background and original image features.

\section{Conclusion}

This paper proposed Subject-Novelty Detection (SND), which disentangles subject and background information and performs novelty detection using subject features only. By reducing the mutual information between subject and background variables and modeling background variation with a deep GMM, SND improves robustness when normal test data come from unseen domains. 

% Experiments on Multi-background MNIST and Kurcuma demonstrate consistent gains over strong novelty detection and domain-shift baselines.

% Future work will explore more complex data (e.g., graphs) and practical deployments, and further study robustness under broader types of domain shifts.

\begin{credits}
\subsubsection{\ackname} The work was partially supported by the General Program of the Natural Science Foundation of Guangdong Province under Grant No.2024A1515011771, the Shenzhen Stability Science Program 2023, and the National Natural Science Foundation of China under Grant No.62376236.
\subsubsection{\discintname}
The authors have no competing interests. 
\end{credits}

\bibliographystyle{splncs04}

% \newpage
\bibliography{pakdd}

\end{document}